\documentclass[twocolumn]{article}

\usepackage{fullpage}
\usepackage[utf8]{inputenc}
\usepackage{graphicx,pgf,tikz,subfig} 
\usepackage{amsmath,amssymb,amsfonts,amsthm} 
\usepackage{multirow,booktabs} 
\usepackage{textcomp} 
\usepackage{enumitem} 
\usepackage{appendix} 
\usepackage{lipsum,mwe} 
\usepackage{algorithm,algpseudocode} 
\usepackage{hyperref,url}

\newcommand{\concat}[2]{\ensuremath{#1\|#2}}
\DeclareMathOperator*{\Concat}{\text{\Large $\|$} }

\newcommand{\reals}{\mathbb{R}}

\usepackage[symbol]{footmisc}

\renewcommand{\thefootnote}{\fnsymbol{footnote}}

\title{Superpixel Image Classification with Graph Attention Networks}
\author{
    Pedro H. C. Avelar\textsuperscript{1,2,*}~\href{https://orcid.org/0000-0002-0347-7002}{\includegraphics[height=10pt]{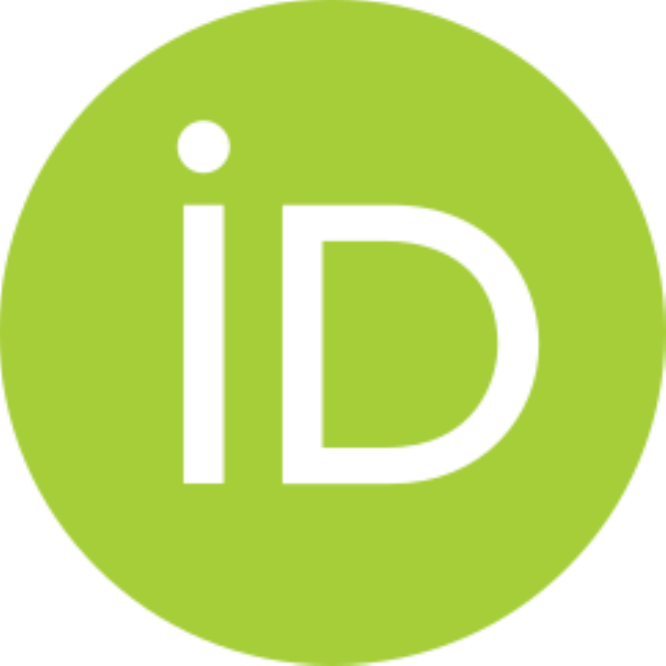}}
    \and Anderson R. Tavares\textsuperscript{1}~\href{https://orcid.org/0000-0002-8530-6468}{\includegraphics[height=10pt]{orcid.pdf}}
    \and Thiago L. T. da Silveira\textsuperscript{3}~\href{https://orcid.org/0000-0001-6788-2667}{\includegraphics[height=10pt]{orcid.pdf}}
    \and Cláudio R. Jung\textsuperscript{1}~\href{https://orcid.org/0000-0002-4711-5783}{\includegraphics[height=10pt]{orcid.pdf}} \quad\quad\noexpand
    \and Luís C. Lamb\textsuperscript{1}~\href{https://orcid.org/0000-0003-1571-165X}{\includegraphics[height=10pt]{orcid.pdf}}\\[1em]
    $1$ Federal University of Rio Grande do Sul, Porto Alegre, Brazil\\
    $2$ Data Science Brigade, Porto Alegre, Brazil\\
    $3$ Federal University of Rio Grande, Rio Grande, Brazil
}

\date{}

\begin{document}

\maketitle

\footnotetext[1]{Corresponding author: \url{phcavelar@inf.ufrgs.br}}
\renewcommand*{\thefootnote}{\arabic{footnote}}
\setcounter{footnote}{0}

\begin{abstract}
This paper presents a methodology for image classification using Graph Neural Network (GNN) models. We transform the input images into region adjacency graphs (RAGs), in which regions are superpixels and edges connect neighboring superpixels. Our experiments suggest that Graph Attention Networks (GATs), which combine graph convolutions with self-attention mechanisms, outperforms other GNN models. Although raw image classifiers perform better than GATs due to information loss during the RAG generation, our methodology opens an interesting avenue of research on deep learning beyond rectangular-gridded images, such as 360-degree field of view panoramas. Traditional convolutional kernels of current state-of-the-art methods cannot handle panoramas, whereas the adapted superpixel algorithms and the resulting region adjacency graphs can naturally feed a GNN, without topology issues.
\end{abstract}

\section{Introduction}

The generic image classification problem consists of determining what object classes (typically from a set of pre-defined categories) are present in an
input image. 
Early approaches followed the traditional pipeline of extracting image features (e.g., colour, texture, etc.) and feeding them to a classifier. The seminal work by Krizhevsky and colleagues~\cite{krizhevsky2012imagenet} explored deep neural networks for image classification, winning the  ImageNet Large Scale Visual Recognition Challenge (ILSVRC) in 2012 by a large margin and setting a turning point for research on image classification. The datasets became more challenging and the networks grew deeper, with the GoogLeNet architecture~\cite{szegedy2015going} winning the ILSVRC2014 challenge and ``Squeeze-and-Excitation'' layers being introduced in~\cite{Hu_2018_CVPR} to win the ILSVRC2017 challenge, with a  top-5 error rate of $2.251\%$.
    
Despite the recent advances both in terms of datasets and network architectures, using traditional convolutional kernels limits the applications of these networks in problems that do not present a domain based on rectangular grids. For example, panoramas capture a full 360-degree field of view. Although the equirectangular representation does use a rectangular domain, sampling is highly non-uniform. To handle these issues, some authors proposed networks designed to adapt to the spherical domain have been designed, such as~\cite{cohen2018spherical}, while others propose to learn how to adapt convolutional layers to the spherical domain, as~\cite{su2019kernel}. As another example, we can mention point-cloud classification, in which spatially unstructured data cannot be represented using a rectangular domain. In this context, some authors either explore a voxelized representation of the scene with 3D networks~\cite{wu20153d} or directly use 3D points as input to a network. 

Graph-based representations can be used to model a variety of problems and domains. Furthermore, they naturally allow several ``multiresolution'' representations of the same object. For example, both pixel-level and superpixel-level representations of the same image might be modeled using graphs. 
In fact, superpixel-based representations have the advantage of reducing the input size, and potentially allowing different domains (e.g. pinhole and spherical images) to be represented as the same (or similar) graph. 
Furthermore, there 
are
several recent advances toward the development of Graph Neural Networks (GNNs)~\cite{wu2020comprehensive}, which could bridge the gap between different domains.
In this paper, we explore Graph Attention Networks (GATs)~\cite{velickovic2018gat} to classify images based on superpixel representations.
GATs are a Graph Neural Network model that combine ideas of graph convolutions \cite{kipf2016semi}, which allows graph nodes to aggregate information from their irregular neighbourhoods, with self-attention mechanisms \cite{Vaswani2017attention}, which allows nodes to learn the relative importance of each neighbour during the aggregation process.


Our methodology comprises the following steps: (i) generate a superpixel representation of the input image; (ii) create a region adjacency graph (RAG) from the superpixel representation, by connecting neighbouring superpixels; (iii) feed the RAG to the GAT, which will predict the class. 
Experiments on several datasets show that the GAT outperforms other RAG-based GNN classifiers, but the RAG provides much less information than the raw image, so that the GAT's performance is inferior compared to the raw-image classifiers.


This paper is organised as follows: in Section~\ref{sec:related}, we present related works and peering approaches. Section~\ref{sec:superpixel} revises existing superpixel segmentation techniques and how they are used to represent graphs, exposing the differences with the competing approach. Section~\ref{sec:model} describes the proposed 
method, while Section~\ref{sec:experiments} shows the experimental 
setup and obtained results. Finally, the conclusions are drawn in Section~\ref{sec:conclusion}.

\section{Related Work}\label{sec:related}

Monti et al. \cite{monti2017monet} provided, to the best of our knowledge, the first application of Graph Neural Networks (GNNs) to image classification, as well as proposing the MoNET framework for dealing with geometric data in general. Their framework
works
by weighting the neighbourhood aggregation through a learnt scaling factor based on geometric distances.

Velickovic et al. \cite{velickovic2018gat} proposed a model using self-attention for weighting the neighbourhood aggregation in GNNs, recognising that this model could be seen as a sub-model of the MoNET framework, nonetheless providing extraordinary results on other datasets, namely Cora and Citeseer, two famous citation networks \cite{sen2008collective}, and on the FAUST humans dataset \cite{bogo2014faust}.

Although graph-based methods can be applied directly to images by considering each pixel a node of the graph, as in the seminal paper of Shi and Malik for image segmentation~\cite{shi2000normalized}, lower-level representations generate smaller graphs. Using each region produced by a segmentation result might be a natural choice,
but generating accurate segmentation results is still an open problem. 
A compromise solution between using individual pixels and object-related regions is \textit{superpixels}. 
Superpixels group pixels similar in colour and other low-level properties, like location, into perceptually meaningful representation units (regions or segments) \cite{stutz2018superpixels}. These oversegmented, simplified, images can be applied in a number of common tasks in computer vision, including depth estimation, segmentation, and object localization \cite{achanta2012slic}. A comprehensive survey on superpixels 
can be found in \cite{stutz2018superpixels}.

The abovementioned work on using GNNs on images, alongside the work on adapting self-attention for GNNs and the works for generating superpixels of images form the pillars on which we based our experiments.

Two other models later came to our knowledge, which extended or could be seen as sub-models of the MoNET framework, using geometric information to weight neighbourhood aggregation, and provided results for the MNIST dataset. One of those is the SplineCNN model \cite{fey2018splinecnn}, which leverages properties of B-spline bases in their neighbourhood aggregation procedure. The other is the Geo-GCN model \cite{spurek2019geogcn}, which is a MoNET sub-model with a differently engineered learned distance function performing data augmentation using rotations and conformations.

Another technique for using GNNs with image data is to use them as a form of semi-supervised augmentation for classification, 
as in \cite{jiang2019data}. The main difference between their method and ours is that while they extract a CNN feature descriptor for each image with a (possibly pretrained) convolutional network, and then build a graph on which their model is used (akin to how Graph Convolutional Networks are used for semi-supervised classification in bag-of-words in \cite{kipf2016semi}), we use the GAT as a classifier for a graph representing an image directly. Although their technique is useful for semi-supervised learning, the technique of using the network for superpixel classification has some possible advantages of its own, and they are not directly comparable.

\section{Superpixel Graphs}\label{sec:superpixel}

A number of techniques exist to generate superpixels from images, such as SLIC \cite{achanta2012slic}, SNIC \cite{achanta2017snic}, SEEDS~\cite{van2012seeds}, ETPS~\cite{yao2015real}, and the hierarchical approach from \cite{wei2018superpixel}. For our experiments, we chose to use SLIC \cite{achanta2012slic} since it was readily available and had a spatial component in its superpixel segmentation. SLIC is stable and it is still recommended among other state-of-the-art oversegmentation algorithms~\cite{stutz2018superpixels}. Nonetheless, we believe that other segmentation techniques with similar characteristics could be used.

\begin{figure*}[ptb]
    \centering
    \subfloat[]{
        \includegraphics[width=.3\linewidth]{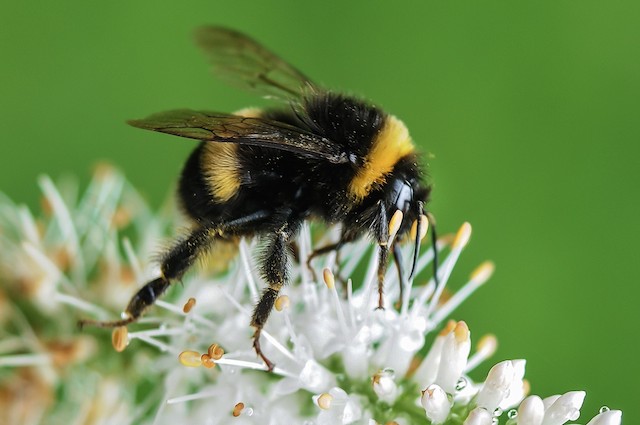}
        \label{fig:superpixelgraph:img}
    }
    \hfil
    \subfloat[]{
        \includegraphics[width=.3\linewidth]{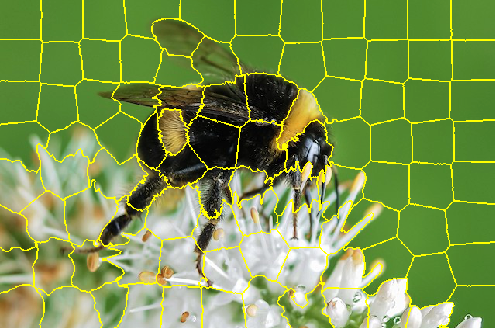}
        \label{fig:superpixelgraph:sp}
    }
    \hfil
    \subfloat[]{
        \includegraphics[width=.3\linewidth]{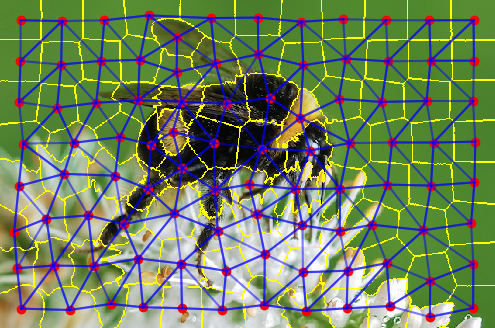}
        \label{fig:superpixelgraph:rag}
    }
    \caption{
        From left to right, the image to be converted into a RAG \protect\subref{fig:superpixelgraph:img},
        the image with the superpixel segmentation being shown \protect\subref{fig:superpixelgraph:sp},
        and the image with the superpixel segmentation and the generated region adjacency graph overlayed on top of it \protect\subref{fig:superpixelgraph:rag}.
    }
    \label{fig:superpixelgraph}
\end{figure*}

After 
applying a superpixel segmentation technique, we generate a Region Adjacency Graph (RAG) by treating each superpixel as a node and adding edges between all directly adjacent superpixels (1-neighborhood connection). 
Note that this differs from the approach adopted in~\cite{monti2017monet}, since their superpixel graphs have connections that span more than one neighbour level, with edges formed with the $K$ nearest neighbours. 
Each graph node can have associated features, providing an aggregate information based on characteristics of the superpixel itself. Algorithm~\ref{algo:superpixelgraph} describes the adopted procedure, whereas Fig.~\ref{fig:superpixelgraph} depicts the generation of a RAG from an image.

\begin{algorithm}[tpb]
\caption{Region Adjacency Graph (RAG) generation}\label{algo:superpixelgraph}
\begin{algorithmic}[1]
\Procedure{Superpixel2Graph}{Image $I$ of width $w$, height $h$ and $k$ channels, Superpixel segmentation technique $S$, and node feature builder $F$}
\State $s, \mathcal{N} \gets S(I)$ \Comment{s returns the superpixel $n \in \mathcal{N}$  of a (x,y) pixel}
\State $x(n) \gets F(I,s,n) \forall n \in \mathcal{N}$ \Comment{$x(n)$ is the feature vector of node n}
\State $\mathcal{E} \gets \{\}$
\For{$1 \leq x \leq w$}
    \For{$1 \leq y \leq h$}
        \If {$s(x,y) \not= s(x+1,y)$}
            \State{ $
                \begin{aligned}
                \mathcal{E} \gets \mathcal{E}\cup\{ & (s(x,y), s(x+1,y))\}
               \end{aligned}$
            }
        \EndIf
        \If {$s(x,y) \not= s(x,y+1)$}
            \State{ $
                \begin{aligned}
                \mathcal{E} \gets \mathcal{E}\cup\{ & (s(x,y), s(x,y+1))\}
               \end{aligned}$
            }
        \EndIf
    \EndFor
\EndFor
\State \textbf{return} $\mathcal{G}=(\mathcal{N}, \mathcal{E}), x$
\EndProcedure
\end{algorithmic}
\end{algorithm}

There are many possibilities for building the features related to each node. For example, statistics about the colour and position of a superpixel, such as the mean, standard deviation, and correlation matrices of its pixels are readily available from the superpixel segmentation. In the case of images defined on rectangular domains, positional information relates to a 2D point. However, this concept can be easily adapted to  omnidirectional images or point clouds, so that a single topology based on graphs can be used in different applications. We do not build features for the edges, since we use an attention-based technique, and believe that the edge feature will be learned accordingly from the attention mechanism using both nodes' features.

In this work in particular, we apply this procedure to the well-known MNIST \cite{lecun1998mnist}, FashionMNIST \cite{xiao2017fashionmnist}, Street View House Numbers (SVHN) \cite{netzer2011svhn} and CIFAR-10 \cite{krizhevsky2009cifar} datasets. The first two datasets contain grayscale images of $28 \times 28$ pixels and 10 classes, and the last two contain RGB images of $32 \times 32$ pixels, both also having 10 classes.

Monti and colleagues~\cite{monti2017monet} used the MNIST dataset and converted it into a graph-based format by using a superpixel-based representation. But whereas they connected nodes through a K-nearest neighbour procedure, we do so using RAGs. Hence, our dataset presents a lower-connectivity graph, which could impair information flow and make the classification problem harder. We also provide results for the RAG representation of the FashionMNIST dataset, since it is a more challenging dataset for which the information loss from the superpixel representation could impact more significantly the model. Since these two datasets contain only grayscale images, we build each superpixel's feature vector as the concatenation of the average luminosity of the pixels in a superpixel and the geometric centroid.

The SVHN and CIFAR datasets, however, both contain RGB channels in their images, and a natural extension for the feature vector is to use the concatenation of the average value for each colour channel and the geometric centroid. These datasets were used to see how the model would perform both with simple and complex colour images.

\section{Our Model}\label{sec:model}

We transform the Undirected Graph produced from the oversegmented image's RAG into a Directed Graph $\mathcal{G}=(\mathcal{N},\mathcal{E})$, and feed it to a Neural Network model that operates on Graphs. More specifically, we use GAT layers stacked on top of each other using the same adjacency graph on each layer.

Our model is a version of the GAT model by Velickovic et al. \cite{velickovic2018gat}, roughly based on the implementation by Nathani et al. \cite{nathani2019kbgat}. Attention is implemented by scattering the source and target nodes' input features into their respective edges, making the transition and activation function on both these inputs and then summing them up over each target node through the edges.

Therefore, for each layer with input dimension $d_i$ and output dimension $d_o$ we learn two functions. The transition function $f : \reals^{2d_i} \rightarrow \reals^{d_o}$, composed of a linear layer followed by a nonlinearity, and the attention function $a : \reals^{2d_i} \rightarrow \reals$ that tells how much the target node of an edge should attend to the source node's information, also composed of a linear layer. 
The values produced by the attention function are activated using softmax for each target node. On the implementation, we take advantage of the fact that $\sigma(\mathbf{z} + \mathbf{c})_{j} = \sigma(\mathbf{z})_{j}$.

\begin{algorithm}[tpb]
\caption{Implemented GAT Layer}\label{algo:gatlayer}
\begin{algorithmic}[1]
\Procedure{GAT-Forward}{Directed graph $\mathcal{G} = (\mathcal{N}, \mathcal{E})$, Node Features $x(n)\forall n \in \mathcal{N}$, learnable transition function $f$ and learnable attention function $a$}
\State $M_{tgt}(t_e,e) \gets \mathrm{1}\{e = (s_e,t_e)\} \forall e \in \mathcal{E}$
\State $h_{src}(s_e) \gets x(s_e) \forall e \in \mathcal{E}$
\State $h_{tgt}(t_e) \gets x(t_e) \forall e \in \mathcal{E}$
\State $h(e) \gets \concat{h_{src}(s_e)}{h_{tgt}(t_e)} \forall e = (s_e,t_e) \in \mathcal{E}$
\State $y(e) \gets f(h(e)) \forall e \in \mathcal{E}$
\State $\alpha(e) \gets a(h(e)) \forall e \in \mathcal{E}$
\State $\alpha_{base}(e) \gets \max_{e \in \mathcal{E}} \alpha(e) $
\State $\alpha_{norm}(e) \gets \alpha(e) - \alpha_{base}(e) \forall e \in \mathcal{E}$
\State $\alpha_{exp}(e) \gets e^{\alpha_{norm}(e)} \forall e \in \mathcal{E}$
\State $\alpha_{sum} \gets (M_{tgt} \times \alpha_{exp}(e)) + \epsilon$
\State $y_{\alpha}(e) \gets y(e)\alpha_{exp}(e) \forall e \in \mathcal{E}$
\State \textbf{return} $o = (M_{tgt}, y_{\alpha}) / \alpha_{sum}$
\EndProcedure
\end{algorithmic}
\end{algorithm}

In summary, given a directed graph $\mathcal{G} = (\mathcal{N}, \mathcal{E})$, with edges $e = (s_e,t_e) \in \mathcal{E}$ and node features $x(n) \forall n \in \mathcal{N}$, let $\mathcal{T}(t)$ be the set of nodes with an edge towards $t$, the attention model can be summarised by Equations \eqref{eq:attention} and \eqref{eq:layer} below, where $\|$ denotes vector/tensor concatenation.

\begin{equation}\label{eq:attention}
    \alpha(s,t) = \frac{e^{a(\concat{x(s)}{x(t)})}} {\sum_{s' \in \mathcal{T}(t)}{ e^{a(\concat{x(s')}{x(t)}) }} }
\end{equation}

\begin{equation}\label{eq:layer}
    o(t) = \sum_{s \in \mathcal{T}(t)} \alpha(s,t)f(\concat{x(s)}{x(t}))
\end{equation}

The detailed algorithm showing the optimisation can be seen in Algorithm~\ref{algo:gatlayer}. 
Each of these layers can be arranged in a multi-head model by concatenating their outputs after the forward pass of each layer. That is, given $k$ heads, the joint output of the $k$-headed layer, where each head has its own transition and attention functions $f_i$ and $a_i$ (as well as the intermediary $\alpha_i$), would be as in Equation~\eqref{eq:multihead}, where $\Concat_{i=1}^k a_k$ is the concatenation of all vectors/tensors $a_k$:

\begin{equation}\label{eq:multihead}
    o(t) = \sum_{s \in \mathcal{T}(t)} \Concat_{i = 1}^k  \alpha_i(s,t)f_i(\concat{x(s)}{x(t)})
\end{equation}

The output of the final GAT layer can then be sum-pooled, having all the values added, and then passed through a MultiLayer Perceptron (MLP) for the final prediction. The Python/Pytorch implementation in its fullest can be seen in the provided GitHub repository\footnote{\url{https://github.com/machine-reasoning-ufrgs/spixel-gat}} as well. Most operations have been parallelized as much as the authors could fathom, with some operations done in a preprocessing phase to avoid overload. 

\section{Experiments}\label{sec:experiments}

In this section, we show 
the potential of our technique on
four datasets: MNIST~\cite{lecun1998mnist}, FashionMNIST~\cite{xiao2017fashionmnist}, SVHN~\cite{netzer2011svhn} and CIFAR-10~\cite{krizhevsky2009cifar}. 
The superpixel algorithm has a target of 75 regions, so that the analyzed graphs present approximately 75 nodes each. 

All experiments were ran either in a computer with a NVIDIA Quadro P6000 or one with a NVIDIA GTX 1070 Mobile. Both computers have 32GB of RAM. For development, we adopted the Pytorch library, version 1.X, using CUDA.

For all experiments we set a budget of 100 epochs for optimisation, with a batch size of 32 images, using a $90/10$ split for training and validation in the dataset's original training data. We use Adam as the optimiser, with a learning rate of $0.001$, $\beta_1 = 0.9$, and $\beta_2 = 0.999$, using the model with the best validation accuracy on the test dataset. If the model failed to leave a baseline accuracy for the first 10 epochs, we restarted the training procedure from scratch.

\subsection{MNIST}\label{ssec:mnist}

We trained two versions of the GAT model: A single-headed GAT with $3$ layers, with $32, 64$ and $64$ neurons, and a two-headed GAT, where each head has the same amount of neurons as the single-headed model. Both models used sum-pooling and a MLP with two layers of $32$ and $d_o$ neurons for the final classification, where $d_o = 10$ is the number of classes in  MNIST. All neurons use ReLU activations, except for those at the last layer of the classification MLP, which use softmax activations. We did not use any regularisation technique.

All dataset images are converted to a corresponding RAG, using SLICO~\cite{slico}, a zero-parameter variant the SLIC algorithm. We set the target number of 
superpixels as 75, but the generated RAGs are not guaranteed to have exactly 75 nodes 
due to how the SLIC algorithm works.

Table~\ref{tab:testresults} shows that both 
GAT models performed better than the MoNET model \cite{monti2017monet}, showing that a learned representation of the geometric distance can lead to better performance than the fixed one of \cite{monti2017monet}. Note that our model has also to deal with graphs that are sparser than the ones used in the baselines~\cite{monti2017monet,fey2018splinecnn,spurek2019geogcn}, since in their graph edges are formed through K-nearest neighbours and ours use only directly adjacent nodes. Although one could expect worse accuracy, the results suggest that our approach is able to learn relevant geometric information relating the features from all neighbouring superpixels. We also report the performance of the graph-based approaches  SplineCNN \cite{fey2018splinecnn} and Geo-GCN \cite{spurek2019geogcn}, as well as recent alternative approaches such as the Generative Tensor Network Classifier (GTNC)~\cite{sun2020generative} and the best performing method based on Support Vector Classification (SVC) as reported in the dataset homepage\footnote{\url{https://github.com/zalandoresearch/fashion-mnist}}. Although our approach does not reach state-of-the-art results, it is competitive and the best among graph-based methods.

\subsection{FashionMNIST}\label{ssec:fashionmnist}

We trained the single-headed and multi-headed GAT models with the same configuration of Section~\ref{ssec:mnist}. Since the FashionMNIST dataset also has $10$ classes, the number of output neurons is $d_o = 10$ as well. 
The RAGs were also generated as in Section~\ref{ssec:mnist}.

Since none of the found graph-based papers present results for both MNIST and FashionMNIST,
we provide 
a 
performance comparison
of
our models
to GTNC~\cite{sun2020generative} and the two best classifiers from the FashionMNIST benchmark \cite{xiao2017fashionmnist} in Table~\ref{tab:testresults}. The gap on the performance between the traditional ML models that  used the full features of the dataset and our models, which use a reduced representation based on the oversegmented image, is higher on FashionMNIST. This shows how much harder the FashionMNIST dataset is for oversegmented images, where the information loss is greater with the aggregation of the features of the pixels in the superpixel.

\begin{table}[tpb]
    \centering
 \caption{Test accuracy for the tested models on MNIST and FashionMNIST, processed as RAGs with approximately 75 nodes (called MNIST-75 and FashionMNIST-75), compared to the baseline models. Bold values show the best of the Graph-based models. We also present the mean accuracies of the two best classifiers for the non-oversegmented MNIST and FashionMNIST datasets, available in the FashionMNIST benchmark.}
       \begin{tabular}{ccc}
    \toprule
                    & MNIST-75  & FashionMNIST-75 \\
    \midrule
        MoNET \cite{monti2017monet}                      & $91.11\%$ & - \\
        SplineCNN \cite{fey2018splinecnn}                  & $95.22\%$ & - \\
        GeoGCN \cite{spurek2019geogcn}                      & $95.95\%$ & - \\
    \midrule
        GAT-1Head                   & $95.83\%$ & $\mathbf{83.07\%}$ \\
        GAT-2Head                   & $\mathbf{96.19\%}$ & $81.40\%$ \\
    \midrule
                    & MNIST     & FashionMNIST \\
    \midrule
        GTNC~\cite{sun2020generative} & $97.6\%$ & $88.2\%$  \\
        SVC(10,poly)                & $97.6\%$* & $89.7\%$* \\
        SVC(100,poly)               & $97.8\%$* & $89.6\%$* \\
    \bottomrule
    \end{tabular}
    \label{tab:testresults}
\end{table}

Another interesting fact to note is that the multi-headed model 
performed worse
than the single-headed one. This was further confirmed when we tried learning with a 4-headed model, which performed slightly worse than the 2-headed one.

\subsection{Street View House Numbers}

\begin{figure}
    \centering
    \includegraphics[width=\linewidth]{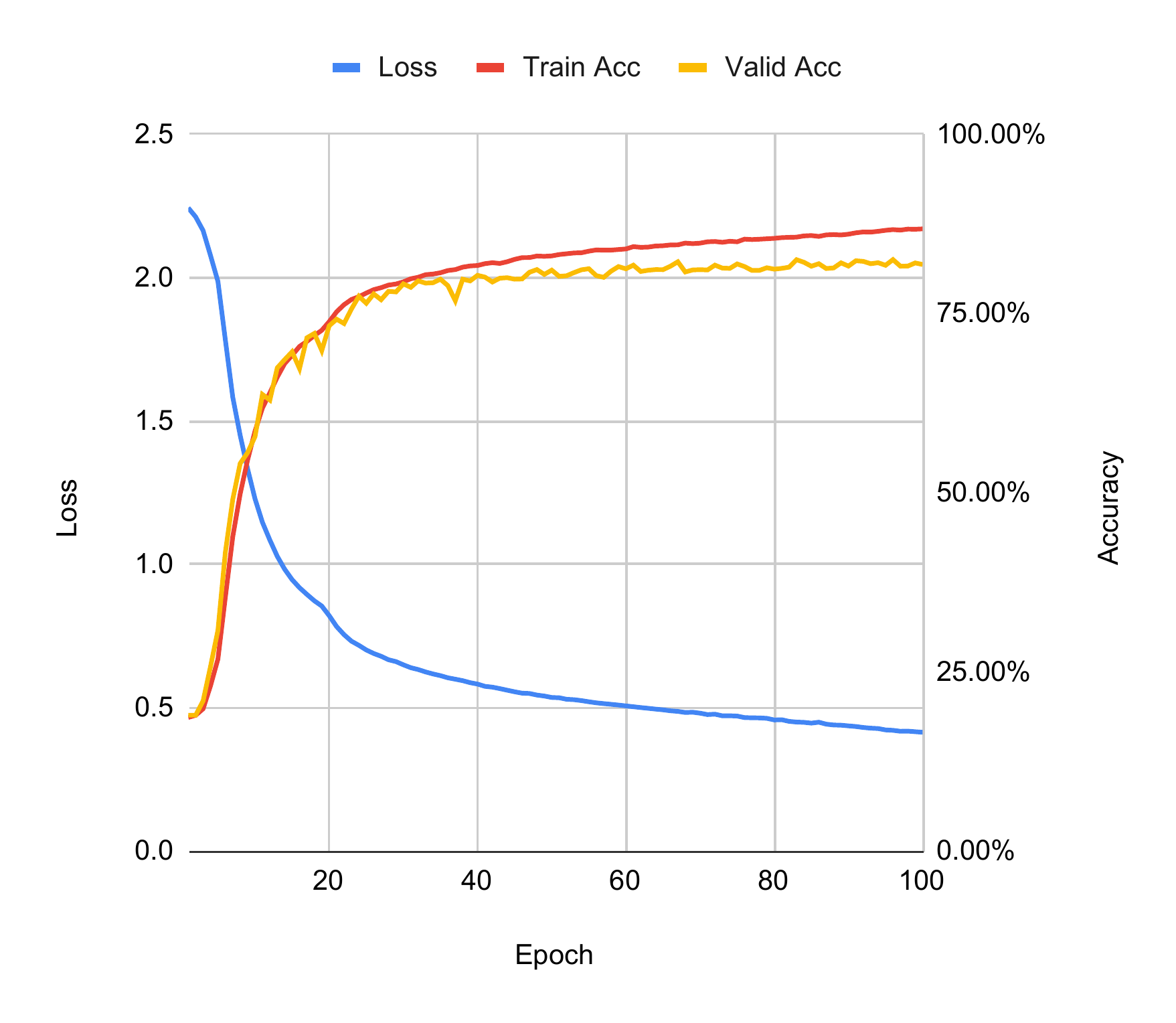}
    \caption{Training Curve for the Street View House Numbers $32\times32$ dataset}
    \label{fig:svhn-training}
\end{figure}

To check the performance of the model with multi-channel data, we trained and tested the two-headed model on the Street View House Numbers (SVHN) dataset~\cite{netzer2011svhn}, using the $32 \times 32$ cropped version and the same parameters for 
oversegmentation
as in the previous sections. We achieved a similar performance with 
the two-headed GAT model on the FashionMNIST dataset, with $ 80.72\%$ test accuracy. This comes to show that our model works even with full colour data, as well as the confounders present in the SVHN dataset. 

That is, the classifier has learned both to prioritise the centermost superpixels and to identify which structure it contains by comparing changes in colour tone, instead of simple changes in luminosity, proving that, although it does not reach state-of-the-art performance, the model can achieve relatively good accuracy even 
working with
less expressive data.


\subsection{CIFAR-10}

The CIFAR-10 dataset~\cite{krizhevsky2009cifar} contains 50,000 $32\times 32$ colour images distributed in 10 classes. We used the same parameters for 
oversegmentation
as in the previous sections.
The results we achieved through a network with the same architecture as GAT-2Head was 45.93\% accuracy on the test set -- very distant from what we achieved on the MNIST and FashionMNIST datasets.
The training and validation accuracies were not impressive either, being 58.61\% and 53.40\% respectively. 

As a baseline, we 
considered
the VGG11 \cite{simonyan2015very} architecture, with and without batch normalisation. We were unable to train the model without
batch normalisation,
whereas with batch normalisation we achieved 62.86\% validation accuracy, which shows the heavy information loss during the RAG transformation procedure. However the comparison of the GAT with the VGG model is still unfair, in terms both of information available to the model and the number of parameters.

\begin{figure}
    \centering
    \begin{tabular}{cccccc}
        \includegraphics[width=.12\textwidth]{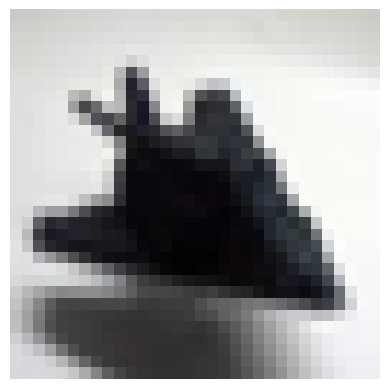}
        & \includegraphics[width=.12\textwidth]{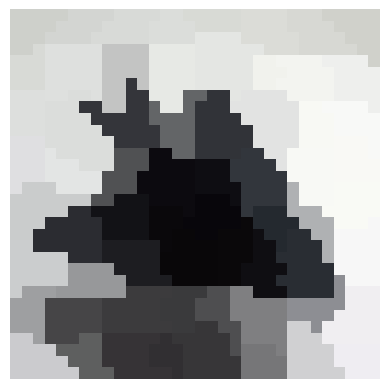}
        & \includegraphics[width=.12\textwidth]{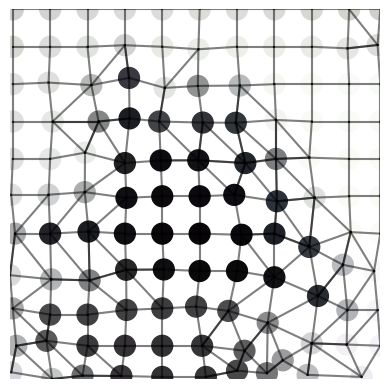}
        \\
        \includegraphics[width=.12\textwidth]{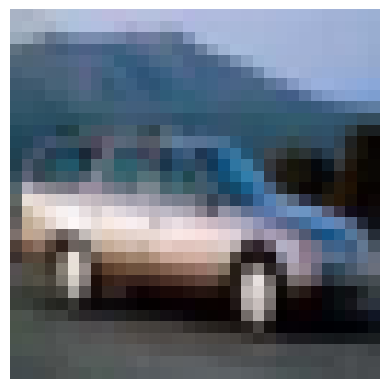}
        & \includegraphics[width=.12\textwidth]{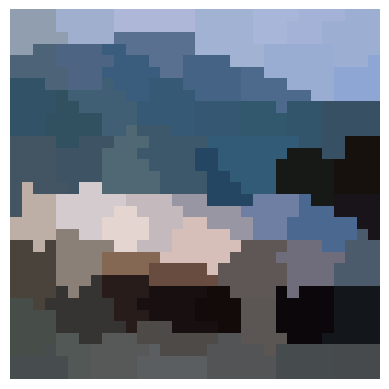}
        & \includegraphics[width=.12\textwidth]{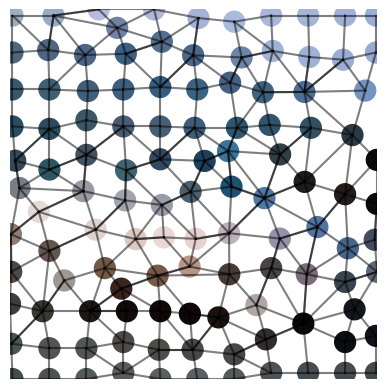}
        \\
        \includegraphics[width=.12\textwidth]{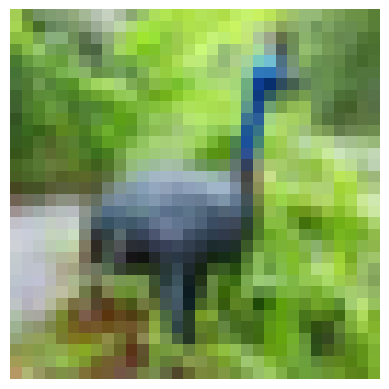}
        & \includegraphics[width=.12\textwidth]{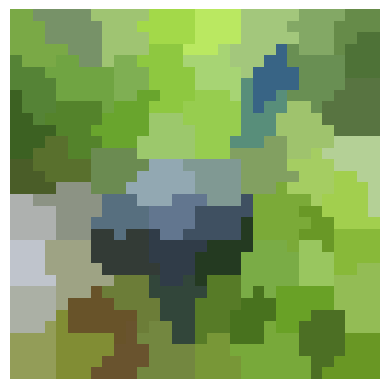}
        & \includegraphics[width=.12\textwidth]{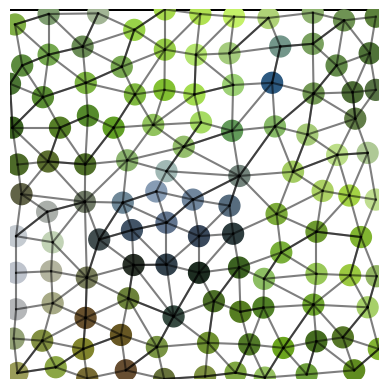}
        \\
        \includegraphics[width=.12\textwidth]{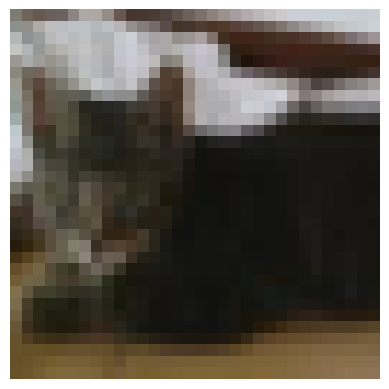}
        & \includegraphics[width=.12\textwidth]{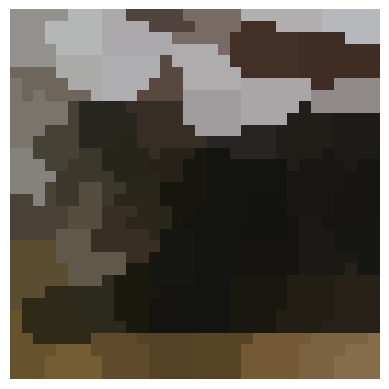}
        & \includegraphics[width=.12\textwidth]{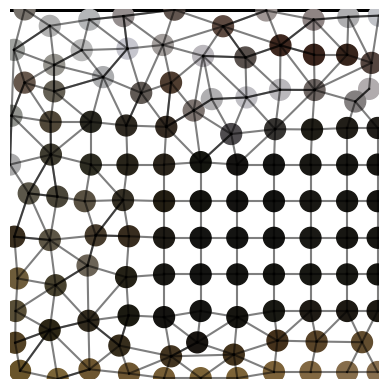}
        \\
        \includegraphics[width=.12\textwidth]{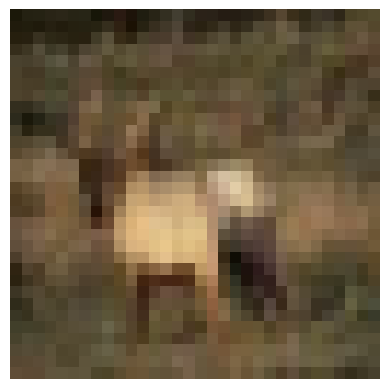}
        & \includegraphics[width=.12\textwidth]{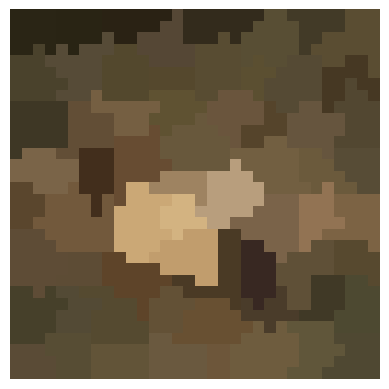}
        & \includegraphics[width=.12\textwidth]{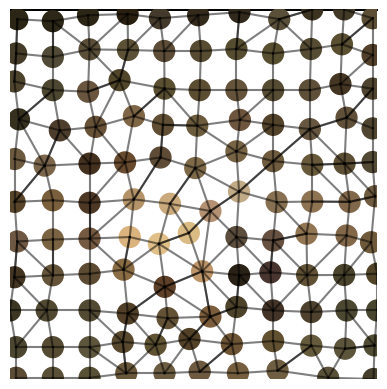}
        \\
        \includegraphics[width=.12\textwidth]{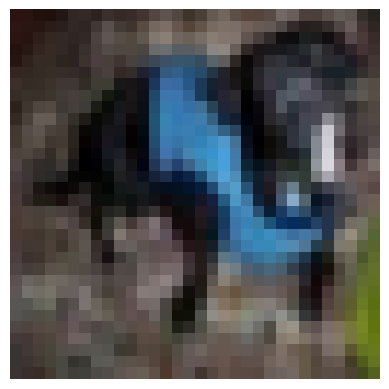}
        & \includegraphics[width=.12\textwidth]{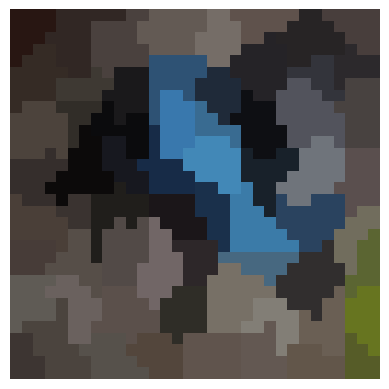}
        & \includegraphics[width=.12\textwidth]{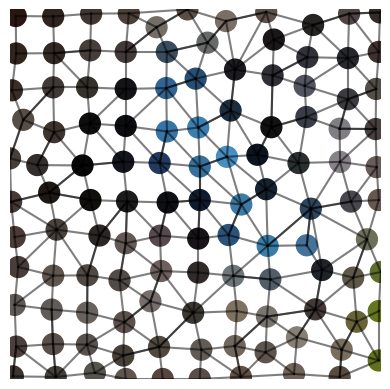}
    \end{tabular}
    
    \caption{Examples showing the loss of information in the RAG procedure when using only the average of each channel, which negatively affects the performance of the network.}
    \label{fig:degradation}
\end{figure}

While the VGG model has access to the oversegmented
image, with each segment's pixel having the averaged RGB values for each superpixel -- approaching also \emph{geometry} information --
the GAT model only has access to the average colour and the centroid position, 
not knowing anything about the superpixel's shape.
More precisely, while the VGG model has access to the middle images in Fig.~\ref{fig:degradation}, the GNN model has only access to the graph on the right images, with each node containing the average pixel value and position.

As for the model size, the VGG11 network has 132,868,840 parameters, while the GAT 
has only 55,364. The VGG network also consumed almost twice as much VRAM as the GAT-2Head architecture on the CIFAR images. 
VGG11 consumed 4,109MiB for training and 1,055MiB for testing, while the GAT model expended 1,067MiB for training and 485MiB for testing. For the sake of illustration, the current state-of-the-art result on CIFAR-10 \cite{huang2019gpipe} is reached with an  AmoebaNet-B (550M parameters) pre-trained on ImageNet and fine-tuned on CIFAR-10. 

\section{Conclusion}\label{sec:conclusion}

In this paper, we have investigated the application and interplay between Graph Attention Networks (GATs) \cite{velickovic2018gat} and  image classification problems. In order to do so, we have used  Region Adjacency Graphs (RAGs) computed from an image segmented using a superpixel algorithm, SLICO.
We showed that using attention-based graph neural networks on a feature space that contains the geometric information can be improved by weighting the edges of a superpixel graph
using a learned function which operates solely on the geometric information.

However, this approach to image classification has its shortcomings. The information loss in the pixel aggregation for more complex images can result in significant performance degradation when compared to using the full image. Also, graph-based approaches may come with the same limitations intrinsic to the models they use, and in our case the GNN-based architecture imposes some limitations in terms of memory usage for larger graphs due to the batching procedure (and thus finer segmentations), despite the smaller number of parameters the model itself had. Training in small batches lead to an unreliable training pattern, further aggravating such issues.

These limitations are, however, venues for future work. It has been shown that architectures based on graph convolutional networks, such as the GAT, suffer from an oversmoothing of node-level information, thus acting like low-pass filters \cite{nt2019lowpassfilters}.
While GATs might not be subject to the same limitation, this could be investigated to allow deeper GATs with potentially better performance in this domain. 
Another venue is helping scaling Graph Neural Networks, of which GAT is a representative, to larger graphs (and thus larger images) or to make them work in an online manner, or with smaller batches.

Also, our models used no regularization whatsoever, and investigating regularization techniques for these models could incur in better performance. Lastly, investigating different node feature vectors could provide the network with richer information and lesser the information loss due to the RAG procedure, possibly with information to recustruct the superpixel's components.

These graph-based approaches to image classification are also a prime example of application to non-euclidean images, such as omnidirectional images \cite{lee2019spherephd}. The flexibility of a graph-based approach could be more invariant to the domain of the image, possibly allowing pre-training on planar images and transfer to spherical images.

\section*{Acknowledgements}We would like to thank NVIDIA Corporation for the Quadro GPU granted to our research group. This work is partly supported by Coordenação de Aperfeiçoamento de Pessoal de Nível Superior (CAPES) -- Finance Code 001 and by the Brazilian Research Council CNPq. We would also like to thank the Pytorch developers for their library.

\bibliographystyle{plain}
\bibliography{superpixelgat}

\end{document}